\documentclass{INTERSPEECH2023}

\interspeechcameraready 

\usepackage{tikz}
\usetikzlibrary{trees}
\usetikzlibrary{automata}
\usetikzlibrary{positioning}
\usetikzlibrary{shapes.multipart}

\title{A Paradigm for Interpreting Metrics and Identifying Critical Errors in Automatic Speech Recognition}
\name{Thibault Bañeras-Roux$^1$, Mickael Rouvier$^2$, Jane Wottawa$^3$, Richard Dufour$^1$}
\address{
  $^1$Nantes University, LS2N, France\\
  $^2$Avignon University, LIA, France \\
  $^3$Le Mans University, LIUM, France}
\email{thibault.roux@univ-nantes.fr, richard.dufour@univ-nantes.fr, jane.wottawa@univ-lemans.fr, michael.rouvier@univ-avignon.fr}

\begin{document}

%author={Ba{\~n}eras-Roux, Thibault and Rouvier, Mickael and Wottawa, Jane and Dufour, Richard},

\maketitle
 
\begin{abstract}
%Automatic Speech Recognition (ASR) research has shown that Word Error Rate (WER) have a poor correlation with human perception in comparison with new evaluation metrics such as SemDist and BERTScore.
%The most commonly used evaluation metrics in automatic speech recognition are WER and CER, both of which are highly criticized for their lack of correlation with human perception.

The most commonly used metrics for evaluating automatic speech transcriptions, namely Word Error Rate (WER) and Character Error Rate (CER), have been heavily criticized for their poor correlation to human perception and their inability to take into account linguistic and semantic information. While metric-based embeddings, seeking to approximate human perception, have been proposed, their scores remain difficult to interpret, unlike WER and CER. In this article, we overcome this problem by proposing a paradigm that consists in incorporating a chosen metric into it in order to obtain an equivalent of the error rate: a Minimum Edit Distance (minED). This approach parallels transcription errors with their human perception, also allowing an original study of the severity of these errors from a human perspective.%percentage. %When this approach is used on words, it is referred to as Minimum Word Edit Distance (minWED), whereas if it is applied to characters, it is known as Minimum Character Edit Distance (minCED).

%The most commonly used evaluation metrics in Automatic Speech Recognition (ASR) are the Word Error Rate (WER) and Character Error Rate (CER). These metrics have been heavily criticized for their lack of correlation with human perception.  Word Error Rate and Character Error Rate have the benefit of being easily interpretable. In this study, we present a paradigm named minWED to keep the interpretability of WER but with scores closer to the human interpretation.
\end{abstract}

\section{Introduction}

% Résumé rapide de l'ASR
Although Automatic Speech Recognition (ASR) performance greatly improved with the recent progress in machine learning and the massive increase in data used for model training, transcription errors are still present, their proportion depending on the context in which these systems are used. 
% Importance de bien évaluer l'ASR pour les humains (comment on l'évalue aujourd'hui ; développement de nouvelles métriques)
%In order to know the capabilities of a system, it needs evaluation. 

Evaluating an ASR system typically involves a comparison between manual ({\it reference}) and automatic ({\it hypothesis}) transcriptions using a chosen metric. For this purpose, the Word Error Rate (WER) and Character Error Rate (CER) are still the most widely used metrics. They consist in calculating a Levenshtein distance, which is an edit distance, between the reference and the hypothesis. However, these measures have been heavily criticized by the community on multiple occasions~\cite{favre2013automatic, ruiz2015phonetically, kafle2017evaluating, gordeeva2021meaning}.
Among the most common identified limits, the WER integrates the same weight for all the words ({\it i.e.} all the words have supposedly the same importance in a sentence), and no linguistic information nor semantic knowledge are taken into account.

\begin{figure}[h!]
    \centering
    \includegraphics[width=0.45\textwidth]{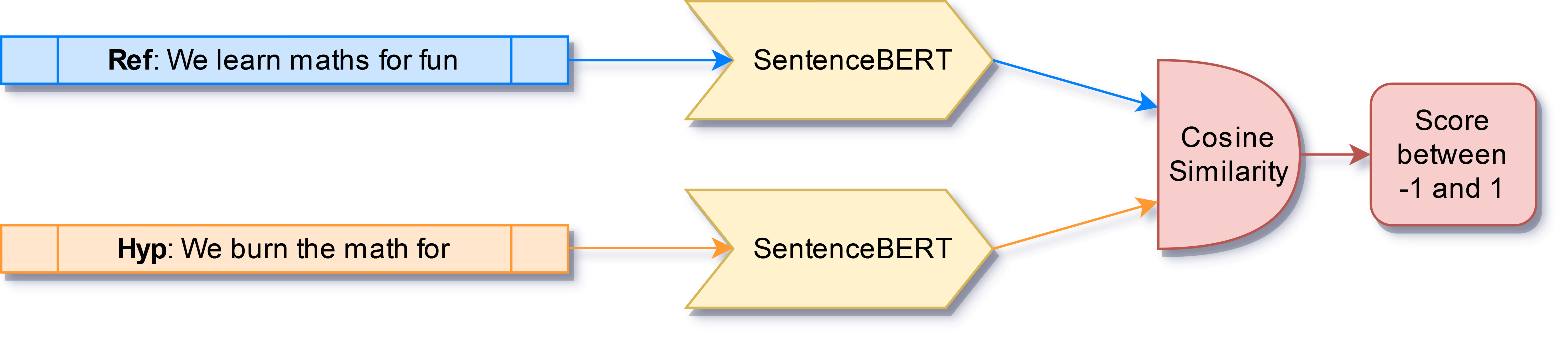}
    \caption{Illustration of SemDist metric.}
    \label{fig:semdist}
\end{figure}

To overcome some of these limitations, embedding-based metrics have been proposed to take into account semantic aspects of transcribed words. 
%which also might be better correlated to human perception.
% Problème d'interprétabilité
Among the propositions, Embedding Error Rate (EmbER)~\cite{roux2022qualitative}, is a WER where the substitution errors are weighted according to the cosine distance between the reference and the hypothesis embeddings at word level obtained from fastText~\cite{grave2018learning, bojanowski2017enriching}. In the same way, \textbf{SemDist}~\cite{kim2021semantic} (See Figure~\ref{fig:semdist}) consists in a cosine similarity distance between the reference and the hypothesis using embeddings obtained at sentence level. \textbf{BERTScore}~\cite{zhang2019bertscore}, applied on various natural language processing (NLP) tasks~\cite{yilmaz2019applying,hanna2021fine}, computes a similarity score for each token in the candidate sentence with each token in the reference sentence using contextual embeddings.

If these metrics allow comparing systems from another view, the obtained score, computed through cosine similarity, is not easily interpretable: while the WER indicates a ratio of correct words on the total, scores of these metrics do not correspond to anything concrete such as words or characters. %scores of these new metrics do not relate to concrete characters or words.
% Objectif du papier (Incorporate SemDist in while still be correlated)

In this paper, we propose to integrate a metric in a novel paradigm, called Minimum Edit Distance (minED), whose purpose is to make embedding-based metric scores interpretable. It is then a question of being able to account for the importance of targeted transcription errors by making it possible to highlight those impacting the most from a human perception point-of-view. This proposal allows us to evaluate and identify the critical errors, which we study in this article in relation to their morpho-syntactic classes, also called part-of-speech or POS.

The paper is organized as follows.  Section~\ref{s:dataset} presents a dataset integrating human perception annotations used to assess the proposed paradigm described in Section~\ref{sec:paradigm}, including its properties, and how it can be used to evaluate error severity. In Section~\ref{sec:analysis}, we will investigate some linguistic aspects as well as the impact of the threshold setting on the correlation performance with human perception. We finally conclude the work and give perspectives in Section~\ref{s:conclusion}.

% Plan du papier et détails des contributions (description d'un nouveau paradigme pour rendre interprétable des métriques d'ASR, évaluation de la corrélation de la perception humaine avec les métriques et les métriques imbriquées dans le paradigme)

%\cite{kim2021semantic} proposed the SemDist metric for ASR evaluation, it consists in projecting the reference sentence and the hypothesis sentence into two sentence embeddings, we then compute a cosine similarity between these two sentences that is meant to represent the semantics (See Figure~\ref{fig:semdist}).

%As said before, the primary goal of this paradigm is to make interpretable an ASR evaluation metric.
%il faut changer la structure du papier.

%In this study, we will work by using the SemDist using Sentence Camembert-large embeddings~\footnote{\url{https://huggingface.co/dangvantuan/sentence-camembert-large}}. 

\section{Dataset with human perception annotations}
\label{s:dataset}
%HATS data set
The HATS dataset\footnote{\url{https://anonymous.4open.science/r/metric-evaluator-82A5/}}~\cite{roux2023hats} is an open-access corpus for French, intended to evaluate the correlation between ASR evaluation metrics and human perception from the reader's perspective. 
It was created using the REPERE corpus~\cite{giraudel2012repere}, containing audio and manually written transcripts of radio and television broadcast in French. The HATS data set was developed with a side-by-side experiment~\cite{gordeeva2021meaning, kafle2017evaluating, kim2021evaluating}: a textual reference was presented to at least 5 subjects, as well as two erroneous hypotheses produced by ASR systems (nine end-to-end systems~\cite{ravanelli2021speechbrain} and two DNN-HMM-based systems~\footnote{\url{https://github.com/kaldi-asr/kaldi/blob/master/egs/librispeech/s5/}}~\cite{povey2011kaldi} with WER between 13.21\% and 30.94\%). The subject must then choose one hypothesis, the best one of the two. This data set thus contains 900 references, each accompanied by two hypotheses and their number of votes.
%évaluation

By calculating the number of times a metric agrees with human annotations ({\it i.e} the metric indicates the best score for the hypothesis chosen by humans), we can calculate a ratio that corresponds to the correlation between that metric and human evaluation. 
%100%, 70%

For some of the choices, human annotators were unsure, thus, it is necessary to take into account only the cases where all subjects agreed (100\%) or where at least 70\% of the solicited humans selected the same hypothesis (70\%). The 70\% threshold was chosen in order to have consistent annotator agreement even if not all participants answers in the same way~\cite{nowak2010reliable}. 
%regarder les corrects, egal et incorrect.

WER and CER metrics both have a limited granularity, which implies a higher probability that both hypothesis have the same score, inducing an equality. For example, for a reference of three words (if we ignore insertion cases), the possible WER values are limited to the following set:%we can only have a WER of the following set: 
$\frac{0}{3}$, $\frac{1}{3}$, $\frac{2}{3}$, $\frac{3}{3}$. While SemDist is able to produce a continuous score ranging between -1 and 1. %For this reason, when computing the correlation of a metric with human perception, we will also indicate the percentage of case where both hypothesis had the same score.

\begin{table}
\centering
\scalebox{0.85}{
\begin{tabular}{|c|c|c|}
\hline
\textbf{Agreement} & \textbf{100\%} & \textbf{70\%} \\ \hline
Word Error Rate & 62.5\% & 53.6\% \\ \hline
Character Error Rate & 74.8\% & 64.1\% \\ \hline
Embedding Error Rate & 71.1\% & 61.6\% \\ \hline
BERTScore BERT-base-multilingual & 83.7\% & 75.7\% \\ \hline
SemDist Sentence CamemBERT-large & 88.8\% & 78.8\% \\ \hline
\end{tabular}
}
\caption{Correlation of each metric according to their human agreement rate.
%If humans were unsure about which transcription is the best, the annotation is ignored.
}
\label{tab:cert_scores}

\end{table}

%semdist étant le meilleur, on travaille sur lui
Table~\ref{tab:cert_scores} presents the correlation of several evaluation metrics depending on the two human agreement rate considered. Note that in this study, SemDist integrates the Sentence-BERT~\cite{reimers2019sentence} version of CamemBERT~\cite{martin2020camembert}~\footnote{\url{https://huggingface.co/dangvantuan/sentence-camembert-large}}, a French version of BERT~\cite{devlin2018bert}, and BERTScore utilizes a multilingual BERT~\cite{devlin2018bert}. We can observe that the SemDist metric has the highest correlation with human perception. Thus, we will consider focusing on this one for our experiments.

%In our work, the HATS data set will be used to test whether the paradigm causes a loss of correlation with human perception compared to the metric used alone.
In our work, the HATS data set will be used to investigate whether the use of the paradigm results in a reduction of correlation with human perception compared to using the SemDist metric alone.

\section{A Paradigm for Metric Interpretation} 
\label{sec:paradigm}
%0) objectif : rendre interprétable une métrique
%1) incorporer une métrique dans minWED ou minCED
%2) number of edits so the metrics is acceptable
%3) plan de la section

%MinED (Minimum Edit Distance) is a new paradigm that consists in calculating a minimum number of words (minWED) or characters (minCED) to be corrected for a hypothesis to be close enough to the reference.

%The purpose of this paradigm is to make interpretable any metrics whose score is not easily understandable.
The purpose of this paradigm is to provide interpretability for metrics that have scores that are difficult to comprehend.
This consists in calculating the minimum number of modifications to be applied to the hypothesis so that it is sufficiently close to the reference regarding its human perception. Following this idea, we apply this method on words (minWED) and on characters (minCED).
%The purpose of this paradigm is to make any of these metrics (\textit{i.e.,} WER, CER, embedding-based metrics) interpretable while remaining correlated to human perception. %(at least, we try).
%Unlike the WER and the CER, the idea is not to compute a score that indicate the number of edits that make the erroneous hypothesis exactly like the reference, the objective is to compute a score that indicates the minimal number of word edits making an erroneous hypothesis acceptable (not necessary error free). %in the light of the hypothesis.
The paradigm is described in Section~\ref{sec:mined}, while Section~\ref{sec:acceptability} defines our measure of acceptability. We then explain different properties of the edit graph (Section~\ref{sec:properties}) and detail the difference between two kind of metrics (consistent, inconsistent), as one of them enables faster computation (Section~\ref{sec:consistent}).

\subsection{Minimum Edit Distance (minED)}
\label{sec:mined}

Word correction involves editing the hypothesis so that there are no more substitutions, insertions, or deletions.
%The main objective of this paradigm is to calculate the minimum number of corrections (word or character) for the hypothesis to be ``acceptable" according to a non-interpretable metric. %For example, a hypothesis with one substitution will need one edit.
The aim of this paradigm is to calculate the minimum number of corrections (words or characters) required to make the hypothesis ``acceptable" according to a non-interpretable metric.
To do so, we can make a graph that represents all the modifications that can be made to the hypothesis so that it becomes the reference (an example can be seen in Figure~\ref{fig:graph}). For each corrected token, we compute a score between the reference and the new hypothesis with the incorporated metric. When the score is below (in the case of a lower-is-better rule) a predefined threshold, the hypothesis is considered ``acceptable". Therefore, we do not need to compute the rest of the graph. It is worth mentioning that it is desirable that the hypothesis be acceptable for humans as well. The score for this pair of reference/hypothesis is the level where the metric is acceptable, which corresponds to the minimum number of edits and leaves some errors in the hypothesis.

Defining the threshold is a hard task that can drastically change results. In Section~\ref{sec:acceptability}, we will explain how the threshold can be defined.

%minWED est un pourcentage d'erreur grave
%WER - minWED est le pourcentage d'erreurs graves

\definecolor{CustomColor}{RGB}{0,225,0} %green

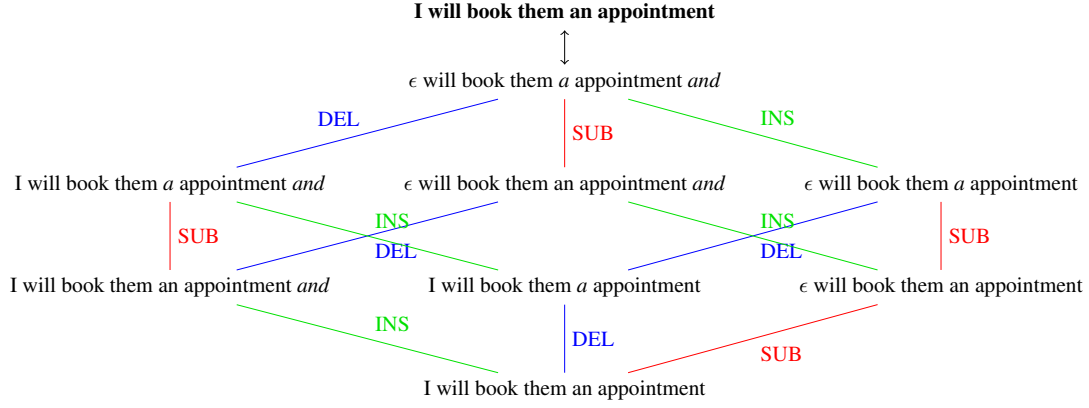
\begin{figure*}
\centering
\scalebox{0.9}{
\begin{tikzpicture}[auto,node distance=25mm, every text node part/.style={align=center}]
\node[](REF) {\textbf{I will book them an appointment}};
\node[](UP)[below=0.5 of REF] {$\epsilon$ will book them \textit{a} appointment \textit{and}};

%---level-1
\node[](L)[below left=1cm and 1cm of UP] {I will book them \textit{a} appointment \textit{and}};
\node[](B)[below=1cm and 0cm of UP] {$\epsilon$ will book them an appointment \textit{and}};
\node[](R)[below right=1cm and 1cm of UP] {$\epsilon$ will book them \textit{a} appointment};

%---level-2
\node[](LL)[below =1cm and -3cm of L] {I will book them an appointment \textit{and}};
\node[](BB)[below =1cm and 0cm of B] {I will book them \textit{a} appointment};
\node[](RR)[below =1cm and -2cm of R] {$\epsilon$ will book them an appointment};
%\node[](BL)[below left=1cm and -3cm of B] {I will book them an appointment \textit{and}\\0.5};
%\node[](BR)[below right=1cm and -3cm of B] {$\epsilon$ will book them an appointment\\0.5};
%\node[](RR)[below right=1cm and -2cm of R] {I will book them \textit{a} appointment\\0.5};

%---level-3
\node[](BELOW)[below =1cm and -2cm of BB] {I will book them an appointment};

\path [<->] (REF) edge        node {}(UP);
\path [blue] % deletion
      (UP) edge [swap] node {DEL}(L)      
      (B) edge        node {DEL}(LL)
      (R) edge        node {DEL}(BB)
      (BB) edge        node {DEL}(BELOW);
\path [red] % substitution
      (UP) edge        node {SUB}(B)
      (L) edge        node {SUB}(LL)
      (R) edge        node {SUB}(RR)
      (RR) edge        node {SUB}(BELOW);
\path [CustomColor] % insertion
      (UP) edge        node {INS}(R)
      (L) edge        node {INS}(BB)
      (B) edge        node {INS}(RR)
      (LL) edge        node {INS}(BELOW);
%\draw[red] (current bounding box.north east) -- (current bounding box.north west) -- (current bounding box.south west) -- (current bounding box.south east) -- cycle; 
\end{tikzpicture}
}
\caption{Computed graph of each possible modification to an error-free hypothesis with the minWED paradigm. Each edge correspond to a corrected error. Given the reference, we have three word errors, each one of a different type: 1 substitution, 1 insertion, 1 deletion. The metric is based on a lower-is-better rule. The token $\epsilon$ correspond to deletions. }
\label{fig:graph}
\end{figure*}

%\subsection{The definition of Acceptability and how to set the threshold}
\subsection{Setting the threshold of acceptability} %Setting the acceptability threshold/
\label{sec:acceptability}

%0) what is acceptability
As discussed in Section~\ref{sec:mined}, the minED indicates the number of edits to make the corrected hypothesis ``acceptable". This notion is based on the idea that there is a value for the metric that is considered as acceptable for humans ({\it e.g} when a human reads an erroneous hypothesis, if a semantic metric indicates a score below the threshold (lower-is-better), the meaning of the original sentence is expected to be understood).

%1) there is no proof of the existence of a threshold of acceptability but we observed in our experiment that some threshold improve or decrease the correlation with human perception (HATS)
To our knowledge, there is no evidence of an universal threshold of acceptability, as it may vary according to context, reader and other parameters. However, we can observe in our experience a tendency towards a threshold that is at least more acceptable than others on average: in Table~\ref{tab:threshold}, we can observe in our experience that some threshold values ($\theta$) do improve the correlation of our paradigm with human perception.
%2) we therefore choose the threshold with the best correlation
If the threshold is set too low, both minWED and minCED metrics will approach WER or CER values (because all edits have to be done in order for the metric score to be below the threshold), and too high values will make them converge towards zero score (because the metric will always be above the threshold and therefore no modification is needed).
%Setting the threshold is difficult as a too small values will make minWED or minCED converge respectively towards WER and CER scores (because all edits have to be done in order for the metric score to be below the threshold), and too high values will make them converge towards zero score (because the metric will always be above the threshold and therefore no modification is needed). 
Our approach was to select a threshold that maximizes the correlation with human perception.

\begin{table}[]
\centering
\begin{tabular}{|c|c|c|}
\hline
\textbf{} & \textbf{100\%} & \textbf{70\%} \\ \hline
SemDist & 88.8\% & 78.8\% \\ \hline\hline
Word Error Rate & 62.5\% & 53.6\% \\ \hline
minWED(SemDist, $\theta$ = 0.004 & 63.6\% & 53.5\% \\ \hline
minWED(SemDist, $\theta$ = 0.024) & \textbf{67.9\%} & \textbf{54.6\%} \\ \hline
minWED(SemDist, $\theta$ = 0.4) & 30.9\% & 22.2\% \\ \hline\hline
%minWED(BERTScore, $\theta$ = 0.072) & 67.34\% & 53.73\% \\ \hline\hline
%minWED(SemDist, $\theta$ = 0.3) & 80\% \\ \hline
%minWED(SemDist, $\theta$ = 0.1) & 70\% \\ \hline
%minWED(SemDist, $\theta$ = 0) & 63\% \\ \hline
Character Error Rate & 74.8\% & 64.1\% \\ \hline
minCED(SemDist, $\theta$ = 0.001) & 77.4\% & 67.1\% \\ \hline
minCED(SemDist, $\theta$ = 0.0095) & \textbf{79.4\%} & \textbf{68.2\%} \\ \hline
minCED(SemDist, $\theta$ = 0.39) & 39.5\% & 29.5\% \\ \hline
%minCER(BERTScore, $\theta$ = 0.0095) & 79.37\% & 68.21\% \\ \hline
\end{tabular}
\caption{Correlation of minWED and minCED metrics according to human agreement rate.}
\label{tab:threshold}
\end{table}

\subsection{Properties of the edit tree}
\label{sec:properties}

The graph is constructed with a node representing the hypothesis produced by the ASR system. If the hypothesis does not contain errors, there is no edit edge and there is one node both for hypothesis and reference. The hypothesis corresponds to the first level and the reference to the last level, and the number of levels is equal to the number of errors + 1.
When there are N errors in the hypothesis, there are N possible edits, so, N nodes in the second level.
As we can see in Figure~\ref{fig:graph}, different edit paths can lead to the same node. If we consider correction as a set ({\it i.e.} it is empty when no corrections are made and full when all corrections have been made), we can consider this graph as an analogy of a Hasse diagram of a graded partially ordered set of a Power set.
Hence, the graph inherits of its properties: 
\begin{itemize}
    \item Number of nodes at level \textit{k} given \textit{n} errors = $\binom{n}{k}$
    \item Total number of nodes given \textit{n} errors = $2^n$
\end{itemize}
%Comment on calcule le nombre de noeuds dans un niveau ?
%Comment on calcule le nombre de noeuds au total dans le graphe ?

%When going from the not corrected hypothesis towards the fully corrected hypothesis, the sum of edges (score of improvement according to metric) is equal to the negative score of the not corrected hypothesis.

%The graph is shaped as a mirror.

%errors     graph shape
%0          {1}
%1          {1,1}
%2          {1,2,1}
%3          {1,3,3,1}
%4          {1,4,6,4,1}
%5          {1,5,10,10,5,1}
%6          {1,6,15,20,15,6,1}
%7          {1,7,21,35,35,21,7,1}
%8          {1,8,28,56,70,56,28,8,1}
%9          {1,9,36,84,126,126,84,36,9,1}
%10         {1,10,45,120,210,252,210,120,45,10,1}
%11         {1,11,55,165,330,462,462,330,165,55,11,1} = 2048

%On observe qu'on peut à chaque fois additionner les deux au dessus (en haut et à gauche), par exemple, 1+1=2, 1+2=3, ..., 5+10=15, 15+20=35

As the complexity of the calculation is exponential, the calculation can be very expensive. For example, for a hypothesis with 5 errors, we have to calculate the metric at most 32. %dans la pratique, il est inutile de calculer le dernier cas puisqu'il est, par définition, égal à la référence.
To overcome this problem, we propose in the Section~\ref{sec:consistent} some optimization solutions.

%For a sentence with $n$ errors, the corresponding graph is composed of $2^n$ nodes. this makes the calculations very expensive.

%E est un ensemble de n éléments, alors l'ensemble des parties de E a 2^n éléments.
%2^0 = 1
%2^1 = 2
%2^3 = 4
%2^4 = 8
%2^5 = 16
%2^6 = 32
%2^7 = 64
%2^8 = 128
%2^9 = 256

%The computational cost can be very high as the metrics need to be computed for each possible edit. In Section~\ref{sec:consistent}, we will talk about the properties of some metrics that can be used to optimize minED calculations.

%\subsection{Difference between ``consistent" and ``inconsistent" metrics}
\subsection{Consistency of metrics}
\label{sec:consistent}
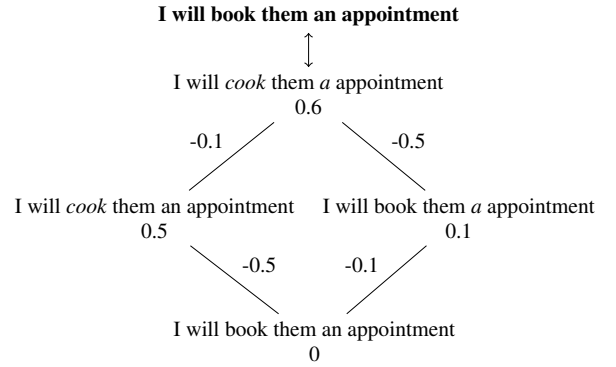
\begin{figure}[htb]
\centering
%\begin{tikzpicture}[->,-latex,shorten >=1pt,auto,node distance=25mm,semithick]
\scalebox{0.9}{
\begin{tikzpicture}[auto,node distance=25mm, every text node part/.style={align=center}]
\node[](REF) {\textbf{I will book them an appointment}};
\node[](1)[below=0.5cm of REF] {I will \textit{cook} them \textit{a} appointment\\0.6};
\node[](2)[below left=1cm and -2cm of 1] {I will \textit{cook} them an appointment\\0.5};
\node[](3)[below right=1cm and -2cm of 1] {I will book them \textit{a} appointment\\0.1};
\node[](4)[below right=1cm and -2cm of 2] {I will book them an appointment\\0};
%\node[](6)[below right of=3]{$110$};
\path [<->] (REF) edge        node {}(1);
\path (1) edge [swap] node {-0.1}(2)
      (1) edge        node {-0.5}(3)
      (2) edge        node {-0.5}(4)
      (3) edge [swap] node {-0.1}(4);
\end{tikzpicture}
}
\caption{Example of impact of correction on consistent metric. Metric is based on a lower-is-better rule.} %il faut trouver un moyen de titrer
\label{fig:consistent}
\end{figure}

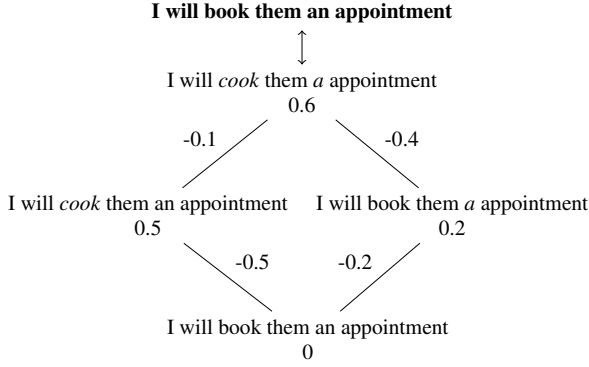
\begin{figure}
\centering
%\begin{tikzpicture}[->,-latex,shorten >=1pt,auto,node distance=25mm,semithick]
\scalebox{0.9}{
\begin{tikzpicture}[auto,node distance=25mm, every text node part/.style={align=center}]
\node[](REF) {\textbf{I will book them an appointment}};
\node[](1)[below=0.5cm of REF] {I will \textit{cook} them \textit{a} appointment\\0.6};
\node[](2)[below left=1cm and -2cm of 1] {I will \textit{cook} them an appointment\\0.5};
\node[](3)[below right=1cm and -2cm of 1] {I will book them \textit{a} appointment\\0.2};
\node[](4)[below right=1cm and -2cm of 2] {I will book them an appointment\\0};
%\node[](6)[below right of=3]{$110$};
\path [<->] (REF) edge        node {}(1);
\path (1) edge [swap] node {-0.1}(2)
      (1) edge        node {-0.4}(3)
      (2) edge        node {-0.5}(4)
      (3) edge [swap] node {-0.2}(4);
\end{tikzpicture}
}
\caption{Example of impact of correction on inconsistent metric. Metric is based on a lower-is-better rule.} %il faut trouver un moyen de titrer
\label{fig:inconsistent}
\end{figure}

When editing a hypothesis in order for it to become like the reference, we can observe an improvement of the score according to the incorporated metric.%~\footnote{It is not impossible that a word correction makes the hypothesis worse with respect to the metric, even if it is unlikely.}.
Correcting can have two known effects: either it improves the score regardless of the previous modifications (see Figure~\ref{fig:consistent}), or it improves the score according to the previous modifications (See Figure~\ref{fig:inconsistent}). The property of consistence is interesting as it allows computing the minimum number of edits faster within the first level. For example, in Figure~\ref{fig:consistent}, correcting the substitution \textit{cook/book} will improve the metric performance of 0.1 no matter if \textit{an/a} was corrected while in Figure~\ref{fig:inconsistent}, correcting \textit{cook/book} will improve the metric performance of 0.5 or 0.4 depending on if \textit{an/a} has been corrected or not.

Therefore, if the incorporated metric is consistent, there is no requirement to compute the entire graph. Instead, a feasible approach is to calculate the second level where a single error in the hypothesis is corrected. 
%Next, substract the score of the original hypothesis with the minimum number of edit improvement so it is below the threshold.
%Next, substract the score of the original hypothesis from the minimum required number of edit improvements so that the resulting score falls below the threshold.
Next, subtract the original hypothesis score from the minimum number of editing improvements required for the resulting score to be below the threshold.
%Next, calculate the minimum number of edits needed to generate a hypothesis with a score below the threshold and add them up.
%Then, sum the minimum amount of edit improvements to have a hypothesis with a score below the threshold.

WER, CER, and Embedding Error Rate are examples of consistent metrics while BERTScore and SemDist are examples of inconsistent metrics.
% Question, est-ce que les métrique embeddings-based de nos jours sont consistent ? Mon avis : EmbER, qui utilise les word embeddings oui, mais SemDist et BERTScore qui utilise des mécanismes d'attention, non.
%In an experiment involving the Sentence Camembert large, we observe a low inconsistency, the standard deviation is below 0.3. %(though, we should seriously check the distribution of scores in order to tell if the inconsistency is important or not.)

\definecolor{classicblue}{RGB}{0,0,255}
\definecolor{cusyellow}{RGB}{255,125,0}

\definecolor{redyel1}{RGB}{255,30,0}
\definecolor{redyel2}{RGB}{255,80,0}
\definecolor{redyel3}{RGB}{255,125,0}

%w\subsection{Measure gravity of errors}
%\textcolor{bitdark}{$\epsilon$} \textcolor{classicblue}{will} \textcolor{red}{cook} \textcolor{classicblue}{them} \textcolor{darkred}{a} \textcolor{classicblue}{appointment} \textcolor{meanred}{and}

%grave:
% cook  (red)
% I/eps (bitdark) -> (redyel1)
% and   (meanred) -> (redyel2)
% a     (darkred) -> (redyel3)

\section{Analysis}
\label{sec:analysis}

\subsection{Linguistic analysis}
Using the minWED paradigm, each error in the hypothesis corresponds either to a word in the reference (substituted or deleted) or to an insertion ({\it i.e.} word only present in the hypothesis). Using a state-of-the-art part-of-speech (POS) tagger for French~\cite{labrak2022antilles}, we propose to linguistically associate a morpho-syntactic class (or POS) to the words in the reference. Our objective is then to analyze the morpho-syntactic classes of errors corrected through the minWED score in order to know if certain classes appear more important in a context of human perception of transcriptions.

\begin{figure}[h!]
    \centering
    \includegraphics[width=0.45\textwidth]{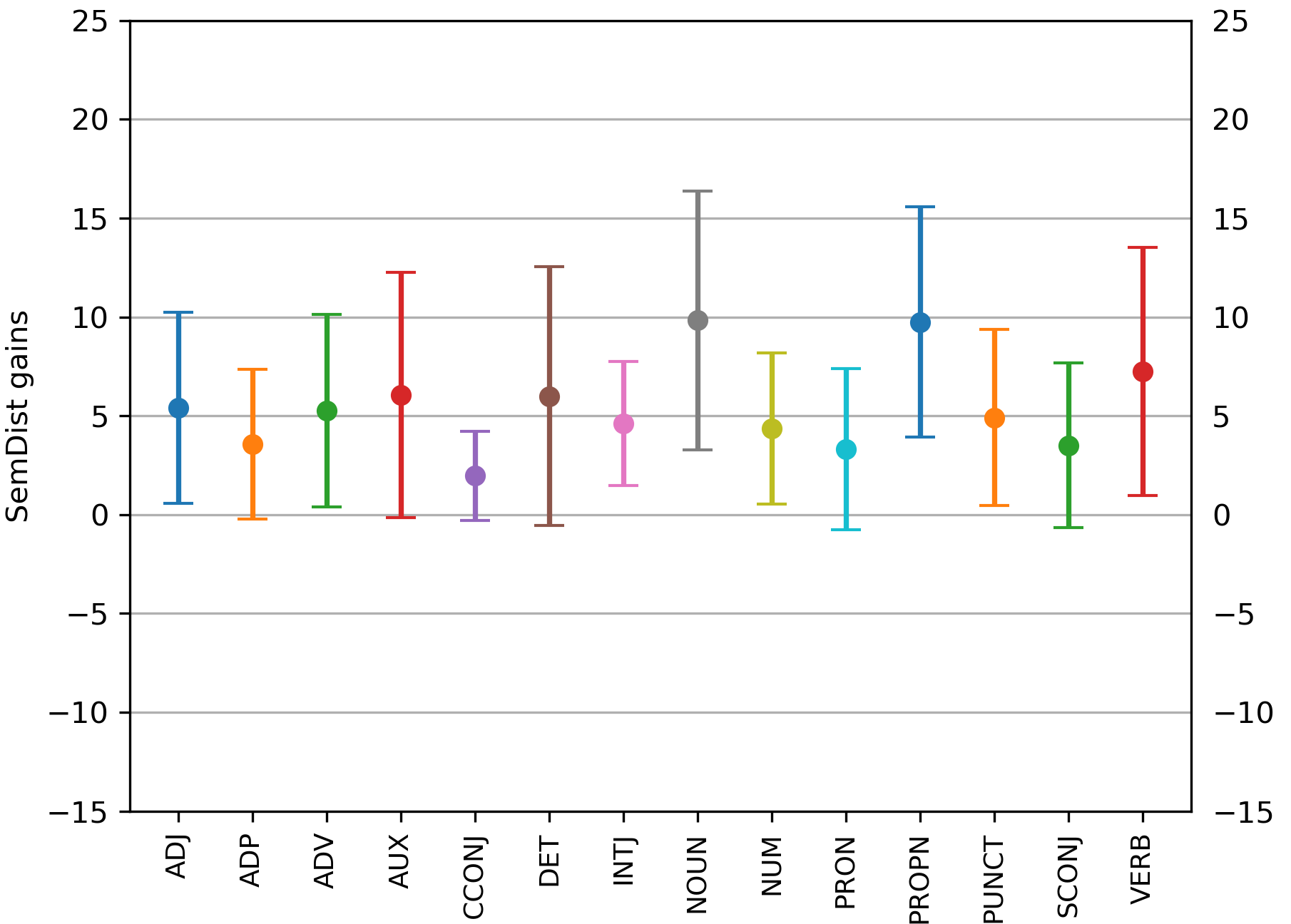}
    \caption{Distribution of SemDist gains for each POS tag corrected. Error bars represent the average gain and the standard deviation. All SemDist values have been multiplied by 100.}
    \label{fig:posgain}
\end{figure}

In Figure~\ref{fig:posgain}, we can observe the distributions of gains per POS tag. 
Across POS, gains are different: the highest gains are obtained for \textit{nouns} or \textit{proper nouns} followed by \textit{verbs}. 
%La phrase suivante va normalement dans la discussion
These three word categories carry important lexical information and are thus crucial for sentence meaning. On the contrary, POS such as \textit{conjunctions} or \textit{pronouns} carry none or only little lexical information. %As a consequence, correcting them will not much impact semantic information of the sentence.
As a consequence, correcting them will not much impact the sentence semantically.%which means that nouns and pronouns correction are the most important factors for semantic improvement of the hypothesis.

Sometimes, correcting at the lexical level decreases the performance according to SemDist and even to human perception. Modifying sentences containing words that are considered wrong because of the unwanted presence or absence of a space, might produce more errors. For example, the hypothesis ``\textit{she worked on state of the art systems}" presents 1 substitution and 3 insertions according to the following reference ``\textit{she worked on state-of-the-art systems}". Correcting the hypothesis might lead to the following: ``\textit{she worked on state-of-the-art of the art}", which appears to be more wrong to human perception but is also a loss according to SemDist.

%the following reference: ``\textit{she worked on state-of-the-art systems}" where the hypothesis is: ``\textit{she worked on state of the art systems}" which is considered as 1 substitution and 3 insertions because of the missing hyphen ; one correction could be: ``\textit{she worked on state-of-the-art of the art}", which appears to be more wrong to our human perception but also for SemDist.

% Un bon exemple de cas où une correction va aggraver l'hypothèse : 
% on voit aujourd'hui où se trouve la grèce
% on voit aujourd' hui où se trouve la grèce -> on voit aujourd'hui hui où se trouve la grèce
%----exemple-en-anglais----
%She worked on state-of-the-art systems
%She worked on state of the art systems -> She worked on state-of-the-art of the art systems

\subsection{Threshold impact}

\begin{figure}[h!]
    \centering
    \includegraphics[width=0.45\textwidth]{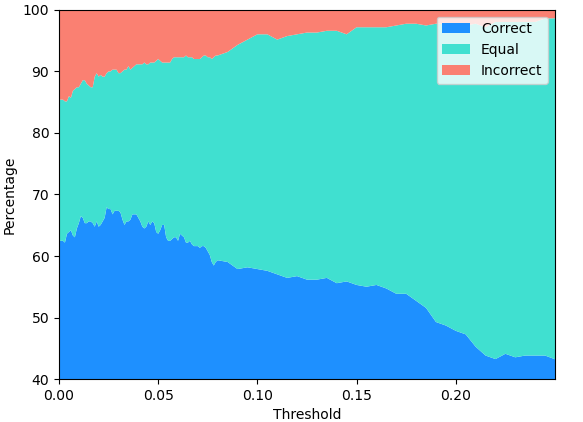}
    \caption{Impact of threshold on minWED predictions on the HATS data set where subjects agreed at 100\%}
    \label{fig:minwed100}
\end{figure}

In Figure~\ref{fig:minwed100}% and Figure~\ref{fig:minwed70}
, we can see the percentage of correct, equal and incorrect minWED predictions on the HATS dataset according to different values of the threshold. When the threshold is close to zero, it is very close to the WER results as minWED goes through the entire graph. When the threshold is increased, the number of correct predictions increases irregularly until it reaches a peak and decreases. We can observe that the higher is the threshold, the more points there are. %The difference of irregularities between low and higher threshold values is due to the fact that we did not calculate with as much precision the values known as bad.
The difference in irregularities between the low and high threshold values is due to the fact that we did not calculate the known bad values as accurately.

%-------------PERCENTAGE--------------

%\section{Scientific questions}

%\begin{itemize}
%    \item Are inconsistent metric that inconsistent? What is the distribution of improvement due to a correction?
%    \item What are the properties of the graph given the number of errors?
%    \item Can we train embeddings to be consistent?
%    \item When using inconsistent embeddings, will an edit move the sentence embeddings toward the same direction? -> Should study the improvement vector.
%    \item Does the minWED evalute the severity of an error or the ease of correcting the error (la gravité d'une erreur ou la facilité à la corriger)
%    \item Les erreurs que minWED ne corrigent pas sont à priori celles que les humains arrivent à ignorer.
%\end{itemize}

%\section{Idées}

%\begin{itemize}
%    \item Comparer les edits que feraient minWED par rapport à différentes métriques (SemDist, Phoneme Error Rate). Soit en visualisant les erreurs comme dans la Section~\ref{sec:visualise}, soit en calculant la différence de façon automatique (rouge foncé - rouge).
%    \item This paradigm can be a tool to analyze errors in ASR.
%\end{itemize}

%section{Discussion}

\section{Conclusions and perspectives}
\label{s:conclusion}

%minED is easier to understand (more interpretable). Unfortunately, it is notable that this method lost an important part of its correlation with human perception. This brings evidence that the Word Error Rate, even when powered with embeddings is not a good indicator of the human perception of errors. On the other hand, this paradigm is interesting to observe what errors are important at the lexical level for humans.

We have proposed a paradigm (minED) allowing both to make Automatic Speech Recognition (ASR) metrics interpretable but also to highlight critical transcription errors from the point of view of human perception. 
Our study indicates that the minED approach presents a more comprehensible strategy for evaluating Automatic Speech Recognition (ASR) systems. Although we observed a significant decrease in correlation with human perception when incorporating a metric in minWED (on words), our findings reveal that minCED (on characters) still exhibits relatively strong performance in capturing error perception. 

These results emphasize the need for a more accurate metric than the traditional Word Error Rate and underline the potential benefits of using minED for identifying critical errors at the lexical level while also promoting a metric that is more closely linked with user perception. 
%while still being interpretable as minED, as it is intended to represent the proportion of important errors.
The study also shows, by the significant loss of correlation to interpretability, that a measure of the number of errors is not the way humans operate. It seems that they prefer to consider the severity of errors rather than the proportion of serious errors.

Another strategy for developing interpretable metrics that correlate with human perception would be to develop qualitative rather than quantitative metrics. For example, the HypRatings dataset~\cite{kim2021evaluating} contains a classical hypothesis and references, but also a qualitative annotation: '\texttt{exact match}', '\texttt{useful hyp}', '\texttt{wrong hyp}', and '\texttt{nonsense hyp}'. It would be worth seeking to develop a metric that would predict these qualitative features.

\section{Limitations}
%\begin{itemize}
%    \item Weak correlation with human perception.
%    \item Consistency of recent metrics.
%    \item Threshold can be dependent of unknown variable. Sometimes, humans will be bore tolerant with some errors. (unsure)
%\end{itemize}
% even if the paradigm improve the interpretation, it decrease the correlation with human perception ; that modern and better metrics are not very consistent and therefore, this paradigm can be very costly ; the acceptability of a correction is very subjective and might not generalize to all humans.

Although the proposed paradigm allows metrics to be interpretable, there are limitations to consider. The use of MinWED and MinED metrics lead to a decrease in correlation with human perception. Depending on the threshold, this decrease may render the use of metrics other than WER irrelevant. Secondly, as modern metrics are not always consistent, minED can be very expensive to compute when there is a high error rate. Finally, it is worth noting that the acceptability of errors is very subjective and might not generalize to all humans.

%\section{Acknowledgements}

%\ifinterspeechfinal
%     The INTERSPEECH 2023 organisers
%\else
%     The authors
%\fi
%would like to thank ISCA and the organising committees of past INTERSPEECH conferences for their help and for kindly providing the previous version of this template.

\newpage

\bibliographystyle{IEEEtran}
\bibliography{mybib}

\end{document}